\pdfoutput=1
\documentclass[11pt]{article}

\usepackage[]{acl}

\usepackage{times}
\usepackage{latexsym}

\usepackage[T1]{fontenc}
\usepackage[utf8]{inputenc}

\usepackage{microtype}

\title{\dataname{}: Commonsense Reasoning for Counterfactual Scene Imagination}

\author{Hyounghun Kim\textsuperscript{\rm *}\;\;\;
Abhay Zala\textsuperscript{\rm *}\;\;\;
Mohit Bansal\\
UNC Chapel Hill \\
\{hyounghk, aszala, mbansal\}@cs.unc.edu}

\usepackage{amsmath}
\usepackage{graphicx, array, blindtext}
\usepackage{multirow}
\usepackage{hhline}
\usepackage{subcaption}
\usepackage{amsfonts,amssymb}
\usepackage{textcomp}
\usepackage{gensymb}
\usepackage{enumitem}
\usepackage{comment}
\usepackage{caption}
\usepackage[switch]{lineno}

\newcommand{\dataname}{\textsc{CoSIm}}

\newcommand\blfootnote[1]{%
  \begingroup
  \renewcommand\thefootnote{}\footnote{#1}%
  \addtocounter{footnote}{-1}%
  \endgroup
}

\begin{document}
\maketitle
\blfootnote{\textsuperscript{*}Equal contribution.}
\begin{abstract}
As humans, we can modify our assumptions about a scene by imagining alternative objects or concepts in our minds. For example, we can easily anticipate the implications of the sun being overcast by rain clouds (e.g., the street will get wet) and accordingly prepare for that. In this paper, we introduce a new task/dataset called Commonsense Reasoning for \textbf{Co}unterfactual \textbf{S}cene \textbf{Im}agination (\dataname{}) which is designed to evaluate the ability of AI systems to reason about scene change imagination. In this task/dataset, models are given an image and an initial question-response pair about the image. Next, a counterfactual imagined scene change (in textual form) is applied, and the model has to predict the new response to the initial question based on this scene change. We collect 3.5K high-quality and challenging data instances, with each instance consisting of an image, a commonsense question with a response, a description of a counterfactual change, a new response to the question, and three distractor responses. Our dataset contains various complex scene change types (such as object addition/removal/state change, event description, environment change, etc.) that require models to imagine many different scenarios and reason about the changed scenes. We present a baseline model based on a vision-language Transformer (i.e., LXMERT) and ablation studies. Through human evaluation, we demonstrate a large human-model performance gap, suggesting room for promising future work on this challenging counterfactual, scene imagination task.\footnote{Our code and dataset are publicly available at: \url{https://github.com/hyounghk/CoSIm}.}
\end{abstract}

\begin{figure*}[t]
    \centering
    \includegraphics[width=1.99\columnwidth]{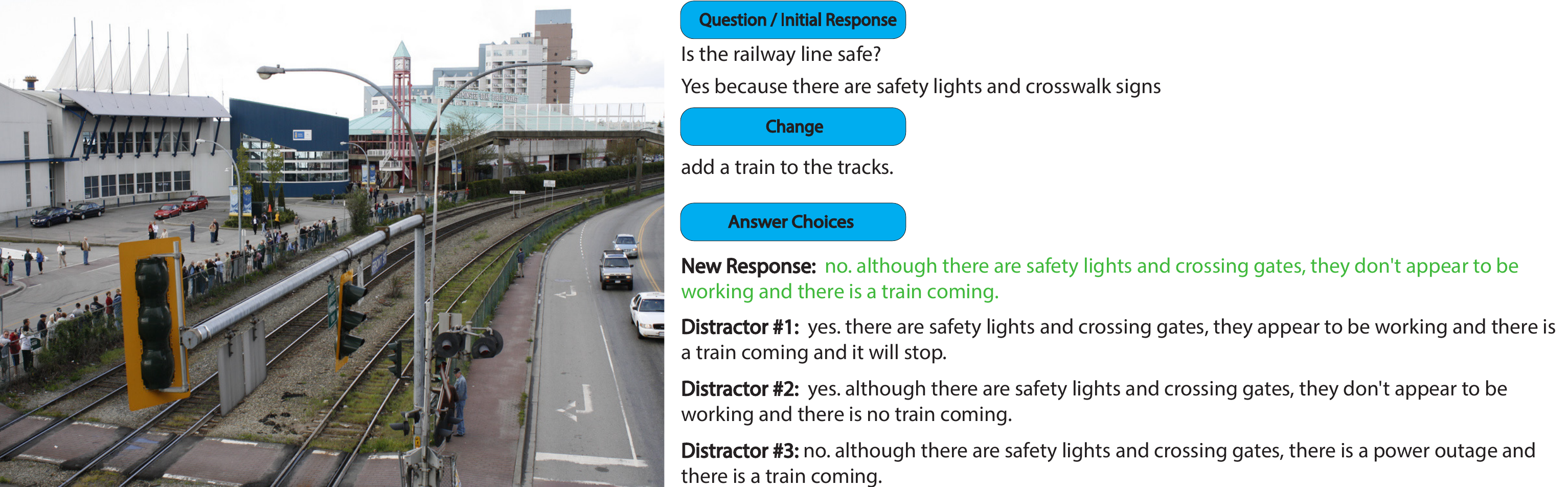}
    \caption{Example from our \dataname{} dataset. An image is associated with an initial commonsense question-response pair, a described counterfactual change to the image, and a new response to the question (randomly shuffled with three human-written distractors).
    \label{fig:main_figure}}
    \vspace{-10pt}
\end{figure*}

\section{Introduction}
Anticipating what would happen when there is a condition change in a situation is an important ability as it allows preparation for the implications of the change. For example, when swimming in the sea on a clear day, you might feel safe. However, if someone told you a storm warning has been issued and dark clouds are coming in soon, you would know that it is no longer safe to swim and return to land.
It will be also very useful to have AI systems that could reason about the implications of such scenario changes and provide appropriate guidance/warnings; however, current AI systems will have a hard time performing such counterfactual commonsense reasoning.

Many efforts have been made to teach machines how to reason about images~\cite{VQA, zhu2016visual7w, johnson2017clevr, hudson2019gqa} and videos~\cite{tapaswi2016movieqa, jang2017tgif, zhu2017uncovering, lei2018tvqa}. This area has been built upon further by efforts to teach machines to use commonsense knowledge when analyzing visual scenes~\cite{pirsiavash2014inferring, wagner2018answering, zellers2019vcr, park2020visualcomet}. Through these efforts, many AI systems have reached near human-level performance on scene understanding tasks. On the other hand, more complex reasoning on scene `changes' has been less explored. \citet{sampat2021clevr_hyp} applies simple condition manipulations (e.g., \emph{``Paint the small green ball with cyan color''}) on synthetic images in a visual question answering setup.
However, this task is based on simple block objects that might not require complex implicit reasoning. Thus, complicated counterfactual commonsense reasoning on scene changes on real-world images and situations remains widely unexplored.

Therefore, in this paper, we introduce a new dataset called Commonsense Reasoning for \textbf{Co}unterfactual \textbf{S}cene \textbf{Im}agination (\dataname{}) to evaluate the commonsense reasoning ability of agents about counterfactual visual scenes imagined via textual descriptions. To be specific, one data instance in our \dataname{} dataset consists of an image, an initial question-response pair, an imagined visual scene change, and a new response with three distractors. The question is about commonsense which can be inferred from the image and the initial response includes a reasoning/justification for its answer. The imagined visual scene change is a textual description of what to modify in the scene to alter the conditions. The new response follows the same format as the initial response, but should be influenced by the imagined change (see Figure~\ref{fig:main_figure}). 

A model for this task needs to take this context information as input and try to predict the correct new response among other distractors. The distractors look similar to the correct new response but have subtle differences and are semantically different from the correct new response, thus a good model on this challenging new multimodal task cannot take shortcuts and needs to fully understand what each choice means based on the context. For example, as shown in Figure~\ref{fig:main_figure}, given an image, the initial question-response pair (\emph{``Is the railway line safe?''} - \emph{``Yes because there are safety lights and crosswalk signs''}), and the scene change (\emph{``add a train to the tracks.''}), models should choose the correct new response (\emph{``no. although there are safety lights and crossing gates, they don't appear to be working and there is a train coming."}) among other distractors (\emph{``no. although there are safety lights and crossing gates, there is a power outage and there is a train coming.''}, etc.). To solve this problem, models need to be able to understand the implications of an incoming train and how safety lights and gates operate at a railroad crossing.

We collect 3.5K high-quality and challenging data instances for this new multimodal reasoning task via a crowd-sourcing annotation platform. To collect each data instance and to help reduce individual crowd-worker load, we break the task up into three separate phases: the question collection phase, the scene change collection phase, and the distractor collection phase. During the distractor collection phase, to help avoid unexpected biases such as text-only, we implement a modified version of Human-And-Model-in-the-Loop Enabled Training (HAMLET) adversarial data collection~\cite{nie2020adversarial} for the validation and test splits. We deploy the model trained on only the textual data and allow annotators to test their distractors against the model as they write (see Figure~\ref{fig:hamlet_figure}).

Our \dataname{} dataset features several diverse types of imagined scene changes (object addition/removal, object state changes, etc.; see Sec.~\ref{sec:change_types} for the full change type list and examples) which requires to deeply understand the contexts, making the task very challenging.
For example, to understand the scene change of \emph{``Add another person to the dock ...''}, the model should figure out what a dock is, where it is located in the image and be able to add one more person onto it via imagination. 

As a baseline model for this new multimodal reasoning task, we employ a vision-language Transformer (based on LXMERT~\cite{tan2019lxmert}) which computes vision and language feature matching scores via multi-head self-attention layers followed by cross-modal attention layers, and we report ablation studies on input modality and scene change types. We also show a large human-model performance gap allowing more effective future work from the community on this new challenging multimodal task on commonsense reasoning for imagined counterfactual scene changes.

\section{Related Work}

\vspace{3pt}
\par
\noindent\textbf{Visual Question Answering.}
There have been many efforts to teach machines how to reason about images~\cite{VQA, zhu2016visual7w, johnson2017clevr, hudson2019gqa} and videos~\cite{tapaswi2016movieqa, jang2017tgif, zhu2017uncovering, lei2018tvqa}, and in some of these tasks, machine performance is approaching human levels. Although these tasks require a complicated reasoning process, they provide very explicit context to solve the problems and might not be enough to evaluate the ability to reason about implicit aspects (i.e., commonsense).  

\vspace{3pt}
\par
\noindent\textbf{Visual Commonsense Reasoning.}
Another actively explored line of study has been on visual commonsense reasoning~\cite{pirsiavash2014inferring, wagner2018answering, zellers2019vcr, park2020visualcomet}. In addition to using the provided clues in the context, these tasks require commonsense knowledge to reason about given problems, making these tasks more challenging since machines should be equipped with prior or external information. However, these tasks handle static scene understanding for which contexts and conditions are not changed during the reasoning process. On the other hand, our proposed \dataname{} introduces an additional dimension of difficulty by integrating imagined scene changes in the context. Moreover, the changes in our \dataname{} dataset are imagined (textually) and counterfactual, so imagination-based commonsense is required for the reasoning. 

\vspace{3pt}
\par
\noindent\textbf{Textual Scene Change.}
Recent effort has been made on visual understanding by requiring mental simulation of changes to the scene~\cite{sampat2021clevr_hyp}. These tasks require simulating change without any visible result, hence increasing the difficulty of VQA tasks. They, however, have been completed in the simpler context of basic shapes and objects and simple questions (E.g. \emph{``How many blue objects will be present in this scene?''}). Our \dataname{} dataset is based on complex real-world images/situations requiring commonsense reasoning about imagined counterfactual scene changes, allowing for evaluation of the ability to anticipate the implications of complex situation changes, thus, future events.

\section{Task}
Given a real-world image, models should predict a new response conditioned on the initial question-response pair and the imagined counterfactual scene change.

\vspace{3pt}
\par
\noindent\textbf{Initial Question and Response.}
The initial question-response pair is created only from a given image. The question and response themselves require quite an amount of commonsense reasoning to understand. For example, to understand the response to the question in Figure~\ref{fig:main_figure} (\emph{``Is the railway line safe?''}), models should know that the \emph{`safety lights'} and \emph{`crosswalk signs'} are devised for keeping people safe around the railway (\emph{``Yes because there are safety lights and crosswalk signs''}).

\vspace{3pt}
\par
\noindent\textbf{Imagined Counterfactual Scene Change.}
The imagined counterfactual scene change is a textual description that modifies the scene in the image. The change affects the reasoning process of the initial question and response, and provides a new context for the new response (\emph{``add a train to the tracks.''}).

\vspace{3pt}
\par
\noindent\textbf{Response on the Scene Change.}
Models should respond to the initial question with a proper reason based on the imagined counterfactual scene change.\footnote{Models should derive scene knowledge from the image or clues embedded in the textual context like the initial response.} The task is a multi-choice setup to pick the correct response among other distractors (\emph{``no. although there are safety lights and crossing gates, they don't appear to be working and there is a train coming.''}). To choose the correct response, models should understand what the implications and safety concerns of an incoming train are and that the safety lights should be turning on and the crossing gates should be closing when a train is in proximity.
\section{Dataset}

\begin{figure}[t]
    \centering
    \includegraphics[width=0.55\columnwidth]{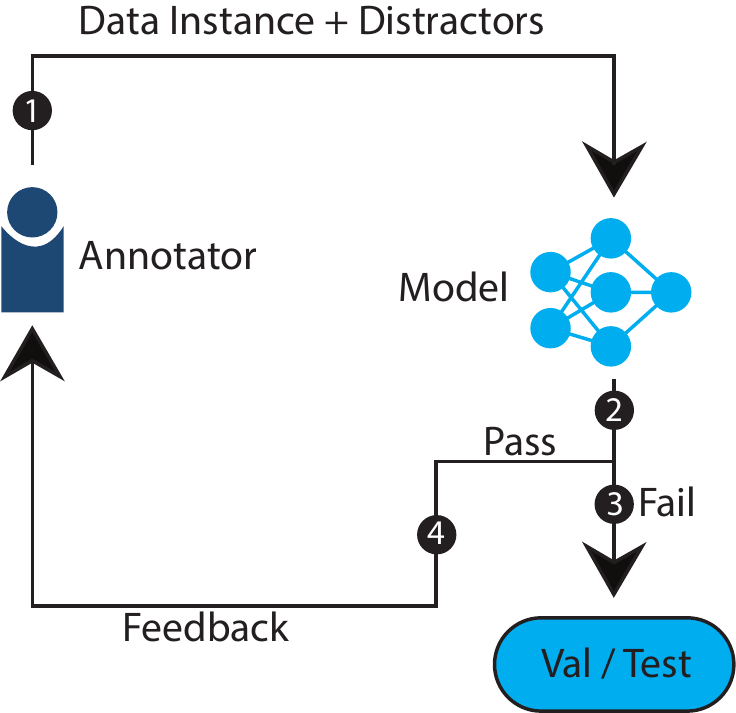}
    \caption{HAMLET cycle for distractor collection on validation and test splits.
    \label{fig:hamlet_figure}}
    \vspace{-10pt}
\end{figure}

Our \dataname{} dataset is composed of 3.5K\footnote{Low resource setup of this middle-scale size dataset encourages employing effective external commonsense knowledge.} images paired with a commonsense question-response pair, a description of an imagined counterfactual change to the image, a new response to the question based on the effect of the described change, and then three distractor responses to the question (all text is in English).

We employ annotators from the crowd-sourcing platform Amazon Mechanical Turk\footnote{https://www.mturk.com}. Our data collection is broken into three separate phases (question, change, and distractor) in order to reduce the workload for each worker. In the question phase, workers are asked to select an image (from three random images) to use, write a commonsense question and then respond to it. In the change phase, they are asked to describe a counterfactual scene change for the image and then write a new response to the initial question. Lastly, in the distractor phase, they are asked to write three distractor responses for the question.

\vspace{3pt}
\par
\noindent\textbf{Commonsense Question Collection.}
To collect the initial question and response, we present three images to the workers and then ask them to choose the one that they want to use (images are taken from Visual Genome~\cite{krishna2017visual}). Then using that image, they should come up with a commonsense question about the image. We define a commonsense question as a question that requires logical thought and understanding of what is happening in the image to be able to answer. Then workers are asked to write a response to their question (the initial response). A response consists of two parts, an ``answer'' that is a direct answer to the question (e.g. \emph{``Yes, ...''}) and then a ``justification'' that uses visual clues from the image to prove the answer is correct (e.g. \emph{``..., because everyone is wearing shorts and short-sleeved shirts and a woman can be seen wearing sunglasses.''}). See Appendix for the collection interface.

\vspace{3pt}
\par
\noindent\textbf{Counterfactual Scene Change Collection.}
In this phase, workers are given the image chosen from the previous commonsense question collection phase and the corresponding initial commonsense question-response pair. Then workers are asked to describe a counterfactual scene change for the image and write a new response to the question based on that scene change (the new response). To help ensure that workers describe a reasonable counterfactual scene change, we provide two guide templates for them to follow when they write. Workers are asked to select the guide template that they believe makes the most sense for them to use for each data instance (see Appendix for collection interface and guide template details).

\vspace{3pt}
\par
\noindent\textbf{Distractor Collection.}
Workers are given the image, the initial commonsense question-response pair, as well as the counterfactual scene change and new response. Then they are told to write three distractor responses that are similar to the new response but incorrect. To help ensure the distractors pose a challenge but are still distinct, we pre-fill the worker's textboxes with the new response. Then they are told to edit the text enough so the answers become false and distinct.

\vspace{3pt}
\par
\noindent\textbf{HAMLET Data Collection.}
To avoid having unexpected biases such as context+response bias in our textual data, when collecting distractors for the validation and test splits, we implement a HAMLET style collection (see Figure~\ref{fig:hamlet_figure}). We deploy the model trained only with textual data and allow workers to test their distractors directly against the model in real-time and check whether they are able to fool it. Workers are also permitted to edit the new response from the previous collection phase if it helps make distractor writing better (they must maintain the original meaning/intent of the new response if they choose to edit).

\vspace{3pt}
\par
\noindent\textbf{Data Verification.}
At each collection phase, we ask workers to verify the previous phase's work. If the previous set of work is not good, workers are given a place to flag and describe the reason for flagging. This reasoning is manually reviewed and if it is fair, then that data is removed and prevented from progressing to the next phase.

\vspace{3pt}
\par
\noindent\textbf{Worker Qualifications and Payment.}
For all 3 phases, workers are required to pass certain qualifications before they could begin. As all of the phases require reading and writing English, they were required to be from native English-speaking countries. Workers were also required to have at least 1,000 approvals from other tasks and a 95\% or higher approval rating. Then for each phase, we require workers to pass a qualification test that tests their understanding of their task at each phase. See Appendix for worker totals and pay (+bonus) rates.

\begin{figure*}[t]
    \centering
    \includegraphics[width=1.9\columnwidth]{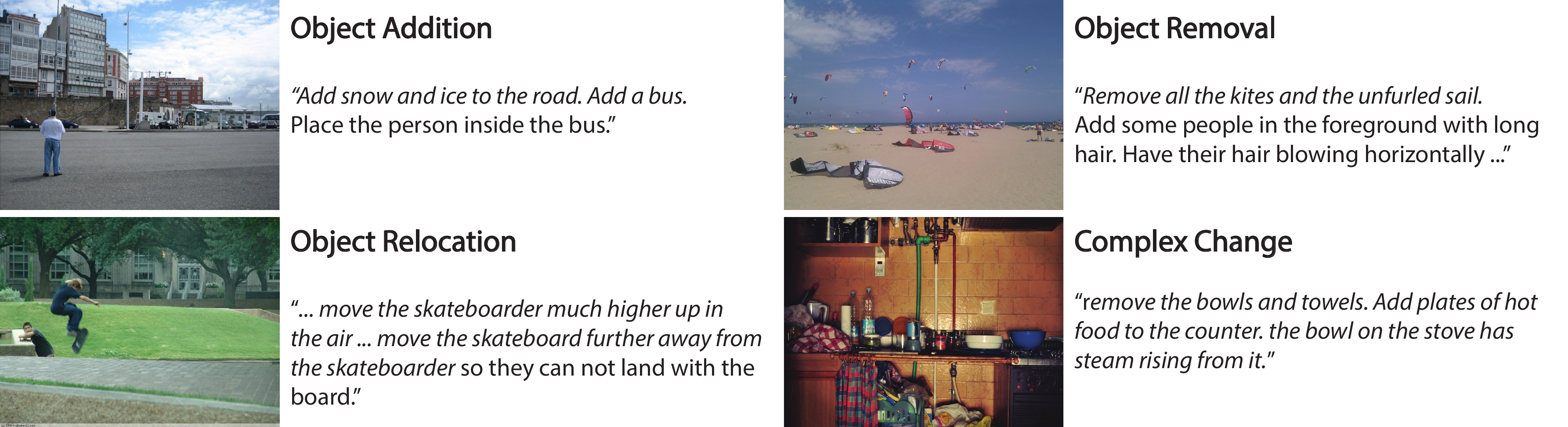}
    \caption{Scene change examples from our \dataname{} dataset. The relevant portions of the change are in italics. Complex changes contain three or more changes within them (this example contains Object Removal, Object Addition, Object State Change).
    \label{fig:scene_change_types}}
\end{figure*}

\section{Data Analysis}
We collect 3.5K task instances (3.5K images, initial questions-response pairs, scene changes, new responses, and 10.5K distractors).

\subsection{Statistics}

\vspace{3pt}
\par
\noindent\textbf{Length.}
Lengths of each part of the data instances are shown in Table~\ref{tbl:lang_stats}. While the lengths of questions are relatively short, the lengths of responses and the changes are long. This means that question itself does not contain detailed clues and models should figure out which information is needed to answer the question. On the other hand, the long responses contain reasons to justify their answers and require models to deeply understand the reasoning process to solve the problem. Furthermore, models should also carefully read the long textual scene change to capture all the condition modifications and apply them to images.

\begin{table}[t]
\begin{center}
\resizebox{0.9\columnwidth}{!}{
 \begin{tabular}{|l|c|c|c|c|}
  \hline

    Component & max. & min. & avg. & sd. \\
    \hline
    \hline
    Question & 22 & 3 & 7.6  & 3.03 \\
    \hline
    Initial Response & 59 & 4 & 18.62 & 8.06 \\
    \hline
    Scene Change & 127 & 3 & 16.08 & 13.01 \\
    \hline 
    New Response & 109 & 5 & 23.38  & 12.40 \\
    \hline 
    Distractor & 111 & 5 & 23.57  & 12.40 \\
    \hline 
\end{tabular}
}
\end{center}
\caption{In our \dataname{} dataset, each part has a different length according to its role and contained information.\label{tbl:lang_stats}
} 
\vspace{-10pt}
\end{table}

\vspace{3pt}
\par
\begin{figure}[t]
    \centering
    \includegraphics[width=0.90\columnwidth]{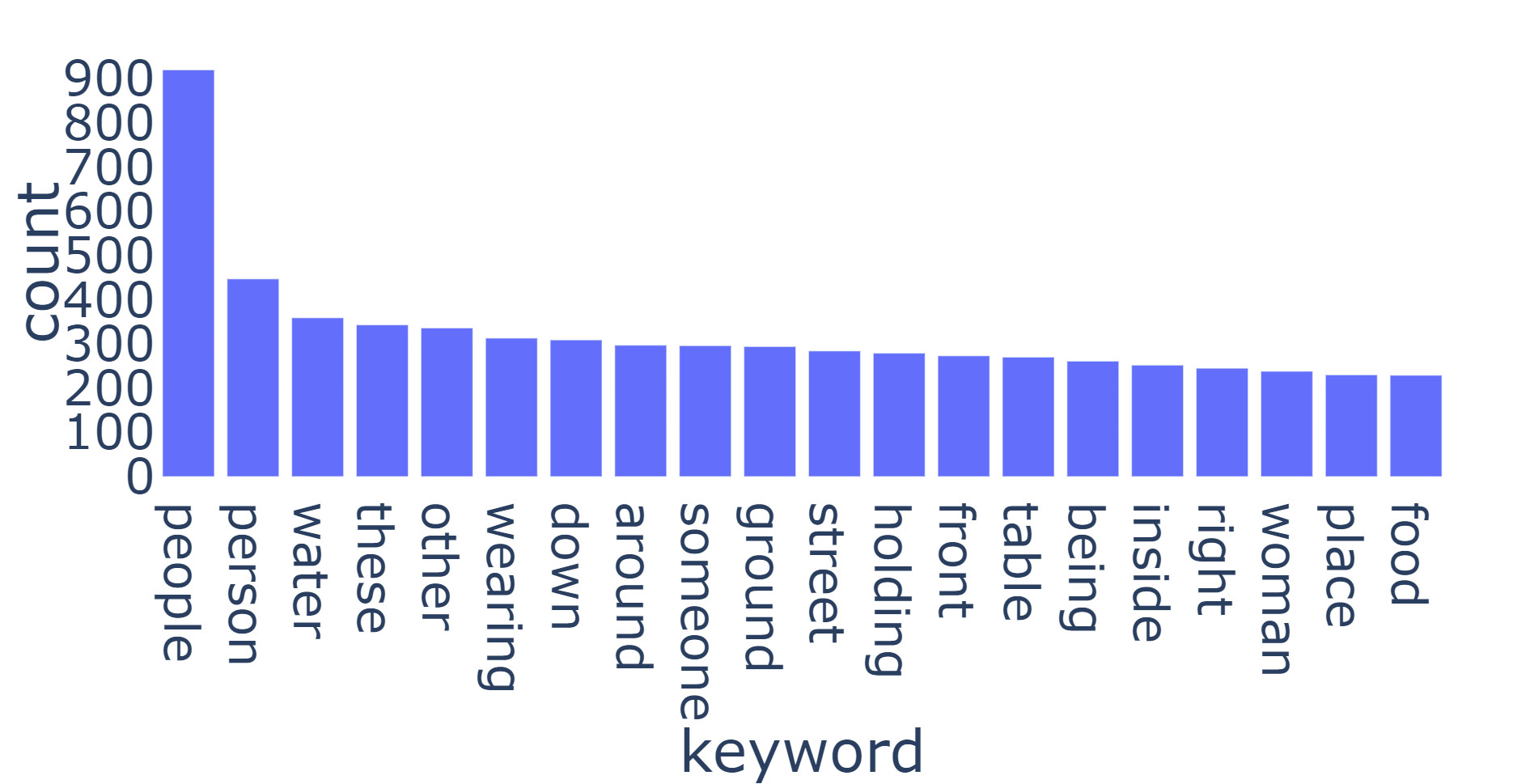}
    \caption{Most commonly occurring keywords in our \dataname{} dataset. Many of them are related to people and directional positioning.
    \label{fig:keywords}}
    \vspace{-10pt}
\end{figure}

\noindent\textbf{Vocabulary.}
Among all data instances in our \dataname{} dataset, there are 9,946 total unique words. Within the commonsense questions, initial responses, scene changes, new responses, and the distractors, there are 3,261 / 4,397 / 4,637 / 5,318 / 6,404 unique words, respectively. The unique word count reflects what is shown by the lengths. Questions are on average the shortest part of each data instance and they have the fewest unique words. The new responses and distractors have long lengths and high unique word counts. The high unique word count for the distractors shows their diversity. Figure~\ref{fig:keywords} shows the most commonly occurring keywords in our dataset. Many of the words are related to people and directional positioning.

\begin{table}[t]
\begin{center}
\resizebox{0.75\columnwidth}{!}{
 \begin{tabular}{|c|c|c|c|c|}

    \hline
    Number of changes present & \multicolumn{4}{c|}{Frequency} \\
    \hline
    \hline
    1  & \multicolumn{4}{c|}{34.70\%} \\
    \hline
    2  & \multicolumn{4}{c|}{35.82\%} \\
    \hline
    3  & \multicolumn{4}{c|}{21.39\%} \\
    \hline
    Greater than 3  & \multicolumn{4}{c|}{8.08\%} \\
    \hline

\end{tabular}
}
\end{center}

\caption{Frequency of number of change types present per instance (from the validation split).\label{tbl:change_types_per_instance_stats}
} 
\vspace{-10pt}
\end{table}

\subsection{Scene Change Type.\label{sec:change_types}}
\begin{figure}[t]
    \centering
    \includegraphics[width=0.90\columnwidth]{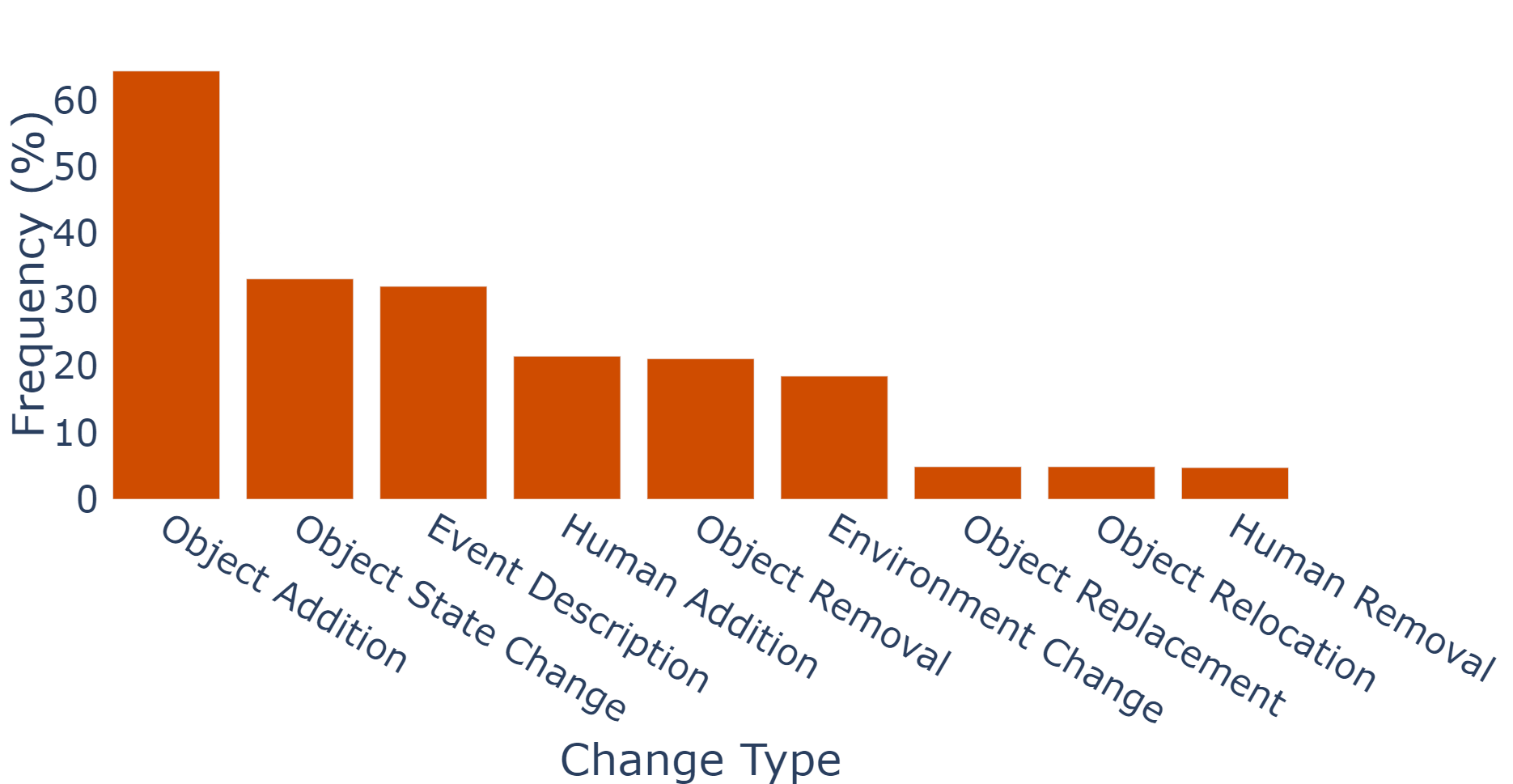}
    \caption{Frequencies of the change types in our \dataname{} dataset (from the validation split).
    \label{fig:change_type_frequencies}}
    \vspace{-10pt}
\end{figure}

Different imagined scene change types are present in our \dataname{} dataset. Imagined scenes changes describe a change (with counterfactual thought) to the image by applying various properties. Some of these scene change types include object addition/removal, object state changes, environment changes, etc. (see Figure~\ref{fig:scene_change_types} for some scene change types and their examples; see Appendix for a figure with a complete list of all the types with examples). These scene change types, while they are seemingly easy to visualize, require a complex understanding of the effect of the change on other elements in the scene. See Figure~\ref{fig:change_type_frequencies} for type frequencies.

\vspace{3pt}
\par
\noindent\textbf{Human/Object Addition.} These two scene change types involve introducing new human(s)/object(s) into the image that was not there prior (\emph{``A bunch of old men are standing next to the birds ...''} / \emph{``There are tears in his eyes ...''}). The object addition scene change type is the most commonly appearing one.
\vspace{3pt}
\par
\noindent\textbf{Human/Object Removal.} These two scene change types involve removing human(s)/object(s) that are visible in the image (\emph{``... remove the workers ...''} / \emph{``Remove the two people's coats''}).
\vspace{3pt}
\par
\noindent\textbf{Object Replacement.} This scene change type involves removing object(s) from the image and replacing them with something else (\emph{``... replace the plates of fruit by plates of dog biscuits ...''}).
\vspace{3pt}
\par
\noindent\textbf{Object Relocation.} This scene change type involves re-positioning object(s). Rather than changing it directly, this type changes its relation to other objects (\emph{``space the zebras out. move them a little further away''}).
\vspace{3pt}
\par
\noindent\textbf{Object State Change.} This scene change type involves altering the state of object(s) present in the image (\emph{``... change her luggage to all have a Burberry pattern ...''}). The alteration of object(s) can occur in various forms such as changing color, size, shape, and orientation (e.g., opening a door).
\vspace{3pt}
\par
\noindent\textbf{Event Description.} This scene change type involves the creation of an event or a description of motion or interaction between objects in the image. This type includes human actions and changes to human emotions (\emph{``A pack of lions are approaching the sheep.''}).
\vspace{3pt}
\par
\noindent\textbf{Environment Change.} This scene change type involves changes that cause large-scale changes to the entire environment either by drastically altering the current environment, creating a new environment, or causing changes in the weather (\emph{``there is very thick dust everywhere''}).
\vspace{3pt}
\par
\noindent\textbf{Complex Changes.} We define a complex change as a change that contains three or more different scene change types. For example, \emph{``someone is throwing snow ball at her''} this change introduces a new human, a new object, and defines an interaction between all these and involves someone already present in the image. These complex changes require much thought to understand their full effect and implications. Complex changes make up about 30\% of our dataset. See Table~\ref{tbl:change_types_per_instance_stats} for change types per instance statistics.

\section{Models}
We employ a vision-language Transformer as the base architecture of our baseline model for the \dataname{} task. To be specific, we use LXMERT~\cite{tan2019lxmert} to compute the score of each context-response pair given an image feature, and select one with the highest score among them.

We employ Faster R-CNN~\cite{ren2015faster} to extract object-level visual features $O=\{o_1, o_2, ..., o_{N_O}\}$ and bound boxes $B=\{b_1, b_2, ..., b_{N_O}\}$ from an image $I$, where ${N_O}$ is the number of detected object features. For textual feature encoding we use BERT~\cite{devlin2019bert} as it is used in LXMERT. We concatenate all the textual data, i.e., question $Q=\{q_1, ..., q_{N_Q}\}$, initial response $R_i=\{r_{i1}, ..., r_{iN_{R_i}}\}$, scene change $C=\{c_1, ..., c_{N_C}\}$, and new response $R_n=\{r_{n1}, ..., r_{nN_{R_n}}\}$ along with [CLS] and [SEP] tokens to create a sequence $W=\{\textrm{[CLS]},Q, \textrm{[SEP]},R_i, \textrm{[SEP]},C,\textrm{[SEP]},R_n,\textrm{[SEP]}\}$ where ${N_Q}$, ${N_{R_i}}$, ${N_C}$, and ${N_{R_n}}$ are the lengths of question, initial response, scene change, and new response, respectively.
\begin{align}
    &O, B = \textrm{FRCNN}(I)\\
    &\hat{O}  = \textrm{Linear}_{O}([\textrm{[V-Tok}_O\textrm{]};O]_{\textrm{dim=t}})\\
    &\hat{B} = \textrm{Linear}_{B}([\textrm{[V-Tok}_B\textrm{]};B]_{\textrm{dim=t}})\\
    &\hat{V}  = \textrm{Linear}_{OB}([\hat{O};\hat{B}]_{\textrm{dim=f}})\\
    &L = \textrm{Emb}(W), \;\; \hat{L} = \textrm{TF}_{L}(L) 
\end{align}
where $\textrm{Linear}_{O}$, $\textrm{Linear}_{B}$, and $\textrm{Linear}_{OB}$ are linear layers. $\textrm{[V-Tok}_O\textrm{]}$ and $\textrm{[V-Tok}_B\textrm{]}$ are visual token attached to object and bounding box sequences (like [CLS] for a language sequence), respectively, and $[;]_{\textrm{dim=t}}$ is concatenation operation along the token-dimension and $[;]_{\textrm{dim=f}}$ is along feature-dimension. $\textrm{TF}_{L}$ is a language Transformer~\cite{vaswani2017attention} which consists of self-attention layers. The $i$th attention head in the $l$th layer $a_{i,l}$ is computed this way:
\begin{align}
    a_{i,l} &= \textrm{Softmax}(\frac{QK^{\top}}{\sqrt{d_{h}}})V\\
    Q=W_{l}^{q}H_{l-1},& \; K=W_{l}^{k}H_{l-1}, \; V=W_{l}^{v}H_{l-1}\\
    H_l &= [a_{0,l}; a_{1,l}; ... ; a_{N_{A},l}]
\end{align}
where $W_{l}^{q}$, $W_{l}^{k}$, and $W_{l}^{v}$ are trainable parameters, $N_{A}$ is the number of attention head, and $d_{h}$ is the dimension of each attention head. Then, $\hat{V}$ and $\hat{L}$ are fed to the cross-attention layers: $\bar{V}, \bar{L} = \textrm{TF}_{X}(\hat{V}, \hat{L})$,
where $\textrm{TF}_{X}$ is cross-attention layers of vision and language Transformer which consists of self-attention layers as well as cross-attention layers. Scores are computed between visual feature and each of the 4 language features (1 ground-truth and 3 distractors) pair: $s_k = \textrm{Linear}(\bar{V}_0 * \bar{L}_{k,0})$, 
where $*$ is the element-wise product, $\hat{V}_0$ is the visual token (i.e., [V-Tok]) that is attached in the input layer, and $\hat{L}_{k,0}$ is the first token (i.e., [CLS]) of $k$-th language feature. The model compares the 4 scores to select the pair with the highest score as the final answer. The loss is computed by cross-entropy: $\mathcal{L} = -\sum_{j}^{N}{\log{p(s_j^{*})}}$, 
where $s_j^{*}$ is a score for the ground-truth pair.

\begin{figure}[t]
    \centering
    \includegraphics[width=0.99\columnwidth]{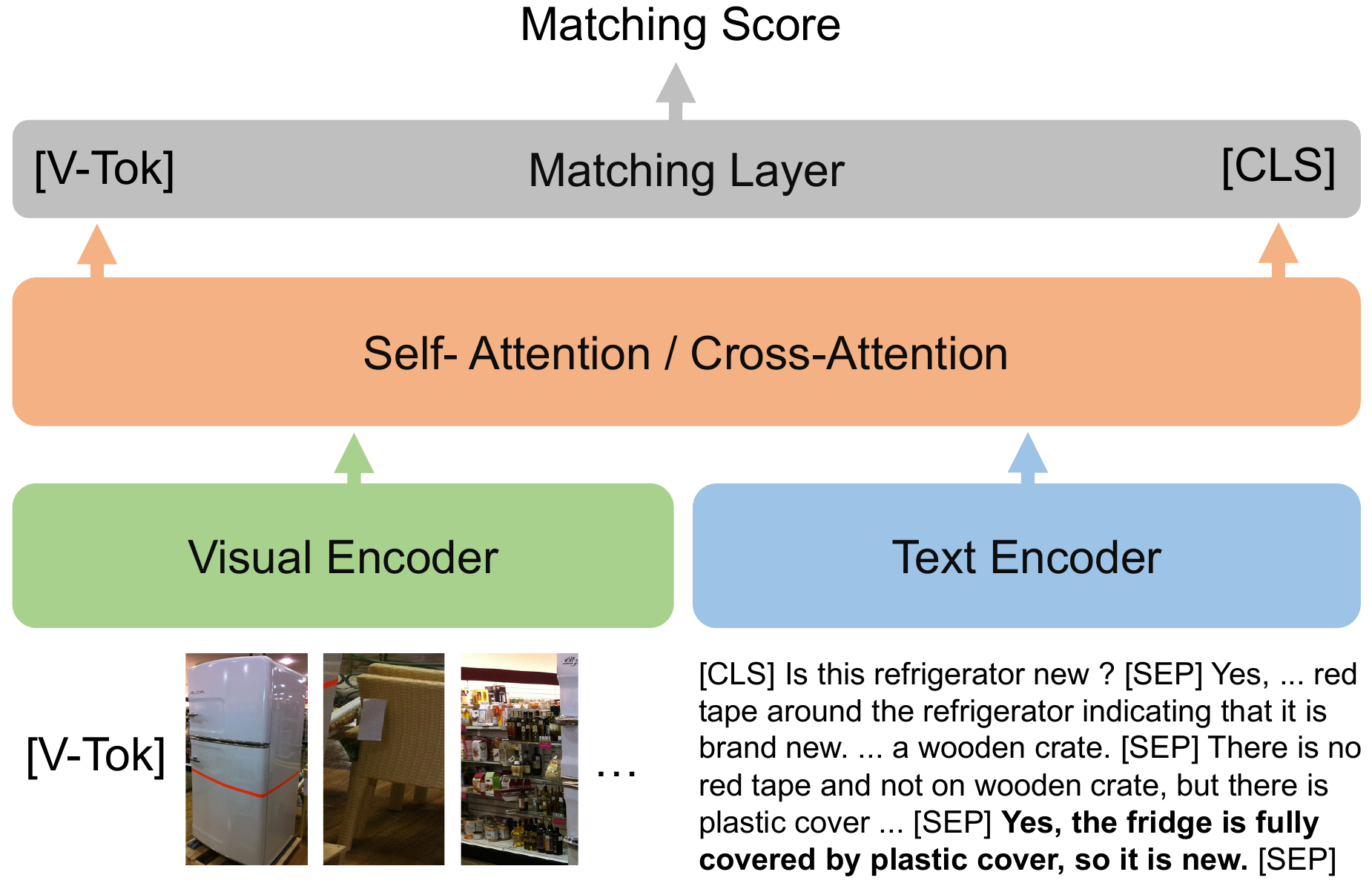}
    \caption{The full model computes the matching scores between [V-Tok] token feature and each [CLS] token feature of the response candidates (a ground-truth response and three distractors), and selects the highest one as a final prediction.
    \label{fig:model_figure}}
    \vspace{-10pt}
\end{figure}
\section{Experiments}

\vspace{3pt}
\par
\noindent\textbf{Data Splits.}
We split the dataset into 1,924/800/800 (train/val/test).

\vspace{3pt}
\par
\noindent\textbf{Training Details.}
We use 768 as the hidden size and use Adam~\cite{kingma2014adam} as the optimizer, setting the learning rate to $1\times 10^{-5}$. See Appendix for more details.

\vspace{3pt}
\par
\noindent\textbf{Human Upper Bound Evaluation Setup.}
We conduct a human evaluation of our \dataname{} task to estimate the upper bound that models can reach. We take 50 samples from the validation split and ask two experts to complete the task and average their scores.

\vspace{3pt}
\par
\noindent\textbf{Scene Change Types.}
We collect the type of the Scene Change for the validation set. Two experts are shown each change and then asked to label it into one or more types. See Figure~\ref{fig:change_type_frequencies} for the change types.

\vspace{3pt}
\par
\noindent\textbf{Multi-Task / Contrastive Learning.}
To exploit extra commonsense reasoning information, we explore multi-task learning (MTL) with a large-scaled visual commonsense reasoning dataset, VCR~\cite{zellers2019vcr} dataset through alternating mini-batch training. In one mini-batch, the model is trained on our \dataname{} dataset, and in the next, the model is trained on the VCR dataset, and so on. Also, we try contrastive learning to explore potential improvement. Specifically, we compute matching scores between each visual token and [CLS] token of each ground truth text feature in a mini-batch, and compute contrastive loss.

\section{Results}
\begin{table}[t]
\begin{center}
\resizebox{0.85\columnwidth}{!}{
\begin{tabular}{|c|c|c|}
  \hline
  
   & Model  & Accuracy (\%) \\
  \hline
  \hline
  1 & Response-Only & 38.37  \\
    \hline
  2 & TC-Response & 44.62 \\
    \hline
  3 & Full (Image-TC-Response) & 49.25 \\
 \hline
  4 & Human & 98 \\
 \hline
\end{tabular}
}
\end{center}
\vspace{-5pt}
\caption{Model results on the val set. Human performance is quite high, showing large room for model improvement (TC: Textual Context). The full model achieves 40.87\% on the test split. \label{tbl:results}
} 
\end{table}

\begin{table}[t]
\begin{center}
\resizebox{0.70\columnwidth}{!}{
\begin{tabular}{|c|c|c|}
  \hline
  
   & Scene Change Type & Accuracy (\%) \\
  \hline
  \hline
  1 & Object Addition & 46.41\\
    \hline
  2 & Object Removal & 37.87\\
    \hline
  3 & Object Replacement & 43.59\\
 \hline
  4 & Object Relocation & 48.72\\
 \hline
 5 & Object State Change & 51.70 \\
 \hline
 6 & Human Addition  & 56.40\\
 \hline
 7 & Human Removal  & 42.10 \\
 \hline
 8 & Environment Change   & 52.70 \\
 \hline
 9 & Event Description   & 51.56 \\
 \hline
\end{tabular}
}
\end{center}
\vspace{-3pt}
\caption{Model performance on different change types. While the model generally shows balanced scores over all scene change types, the performance on removal types seems to be lower than addition types.\label{tbl:results_type}
} 
\vspace{-10pt}
\end{table}

\begin{table}[t]
\begin{center}
\resizebox{0.8\columnwidth}{!}{
\begin{tabular}{|c|c|}
  \hline
  
   Number of Scene Change Types & Accuracy (\%) \\
  \hline
  \hline
  1 &  52.35 \\
    \hline
  2  & 46.34 \\
    \hline
  3 or more & 49.15 \\
 \hline
\end{tabular}
}
\end{center}
\vspace{-3pt}
\caption{Model performance on different numbers of change types, showing instances with single scene change type are relatively easier. \label{tbl:results_type_num}
} 
\vspace{-3pt}
\end{table}

\begin{table}[t]
\begin{center}
\resizebox{0.7\columnwidth}{!}{
\begin{tabular}{|c|c|c|}
  \hline
  
   & Model  & Accuracy (\%) \\
  \hline
  \hline
  1 & MTL with VCR & 47.37 \\
    \hline
  2 & Contrastive Learning & 49.50\\
    \hline
\end{tabular}
}
\end{center}

\caption{Model performance on multi-task and contrastive learning approaches.\label{tbl:results_adv}
} 
\vspace{-10pt}
\end{table}

\paragraph{Modality Ablation.}
We build models with different input modalities and conduct an ablation study. As shown in Table~\ref{tbl:results}, the Response-Only model (which only takes the new response/distractors as input) does not do well (row 1). The TC-Response model (which takes all text data as input) obtains a better score than the Response-Only model (row 1 and 2), but still performs poorly. The Full model (which takes the full image and text data as input) does best (row 3), with a val/test score of 49.25/40.87, meaning models need all the visual and textual input to perform reasonably.\footnote{The standard deviation of the Full model's scores on validation split is 1.52}\textsuperscript{,}\footnote{The average length difference between the predicted responses and the rest is 0.0075 words, and the ground-truth response indices are randomly assigned, thus there is no bias based on the response length and index.}

\vspace{3pt}
\par
\noindent\textbf{Human Evaluation.}
We conduct a human evaluation to check the upper performance bound for the \dataname{} task. As shown in row 4 of Table~\ref{tbl:results}, the score is quite high\footnote{Inter-annotator agreement (kappa) is 0.9461, which indicates nearly perfect agreement.}, indicating a large room for improvement from future work.

\vspace{3pt}
\par
\noindent\textbf{Scene Change Types.}
As shown Table~\ref{tbl:results_type}, our model shows balanced scores over all scene change types in general, however, comparing the addition and removal types (row 1 and 2 for object, row 6 and 7 for human), the performance on removal types is lower than addition types. That is possibly because removing something from an image might be harder to imagine.

\vspace{3pt}
\par
\noindent\textbf{Number of Scene Change Types.}
As shown Table~\ref{tbl:results_type_num}, instances with a single scene change type seem to be relatively easier to address than ones with multiple scene change types. This might imply that multiple scene changes make the reasoning process more complex and challenging. 

\vspace{3pt}
\par
\noindent\textbf{Multi-Task / Contrastive Learning.}
As shown in row 1 of Table~\ref{tbl:results_adv}, multi-task training with VCR does not seem to help improve performance on our \dataname{} dataset, implying our dataset is challenging to address and requires a more complex reasoning process. The performance of the contrastive learning (row 2) is also very close to the Full model's (row 3 in Table~\ref{tbl:results}), meaning more advanced approaches might be needed to tackle our \dataname{} dataset/task.

\vspace{3pt}
\par
\begin{figure}[t]
    \centering
    \includegraphics[width=0.90\columnwidth]{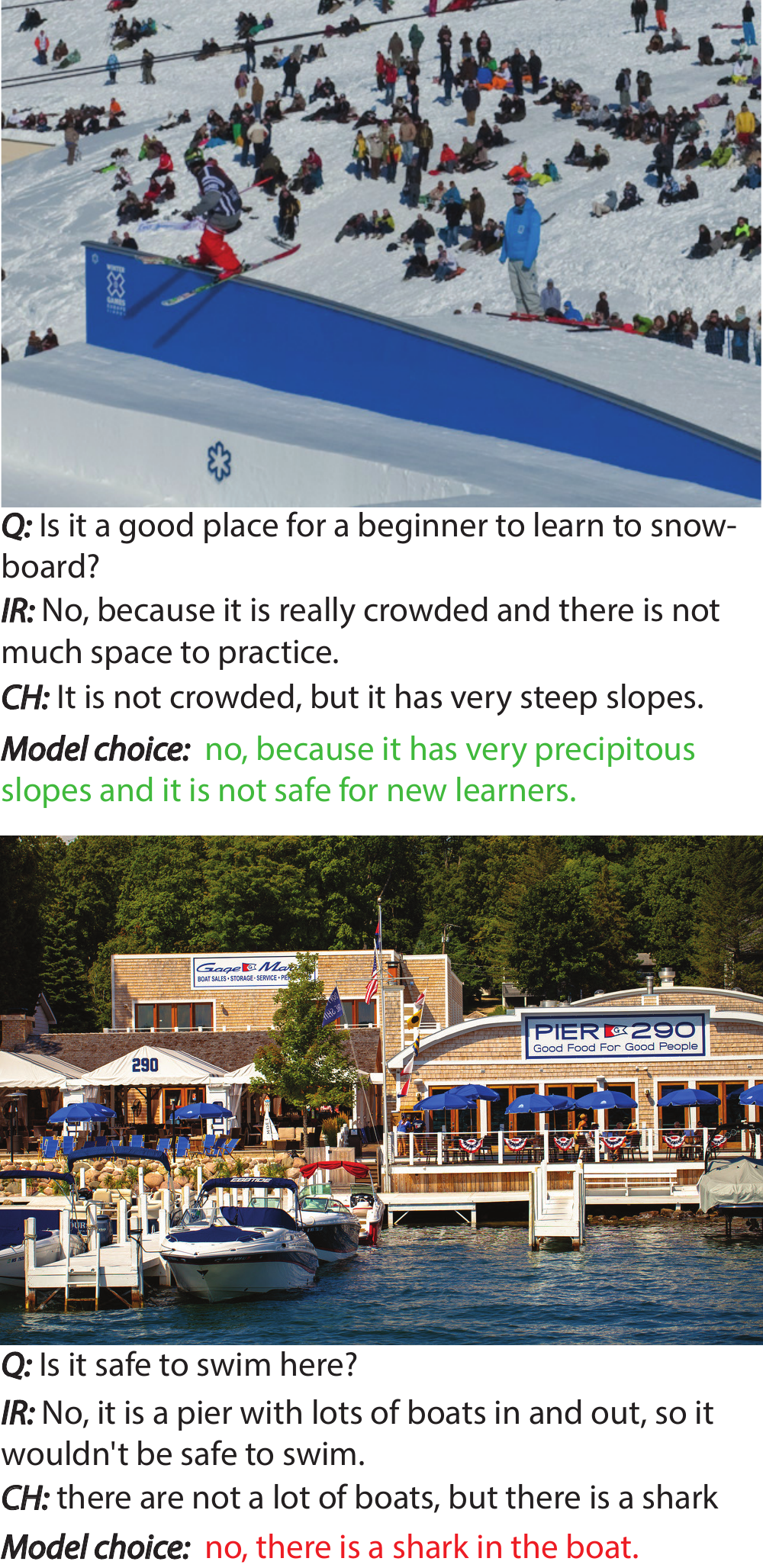}
    \caption{Model output examples (Q: question, IR: initial response, CH: scene change).
    \label{fig:model_outputs}}
    \vspace{-10pt}
\end{figure}
\begin{figure}[t]
    \centering
    \includegraphics[width=0.90\columnwidth]{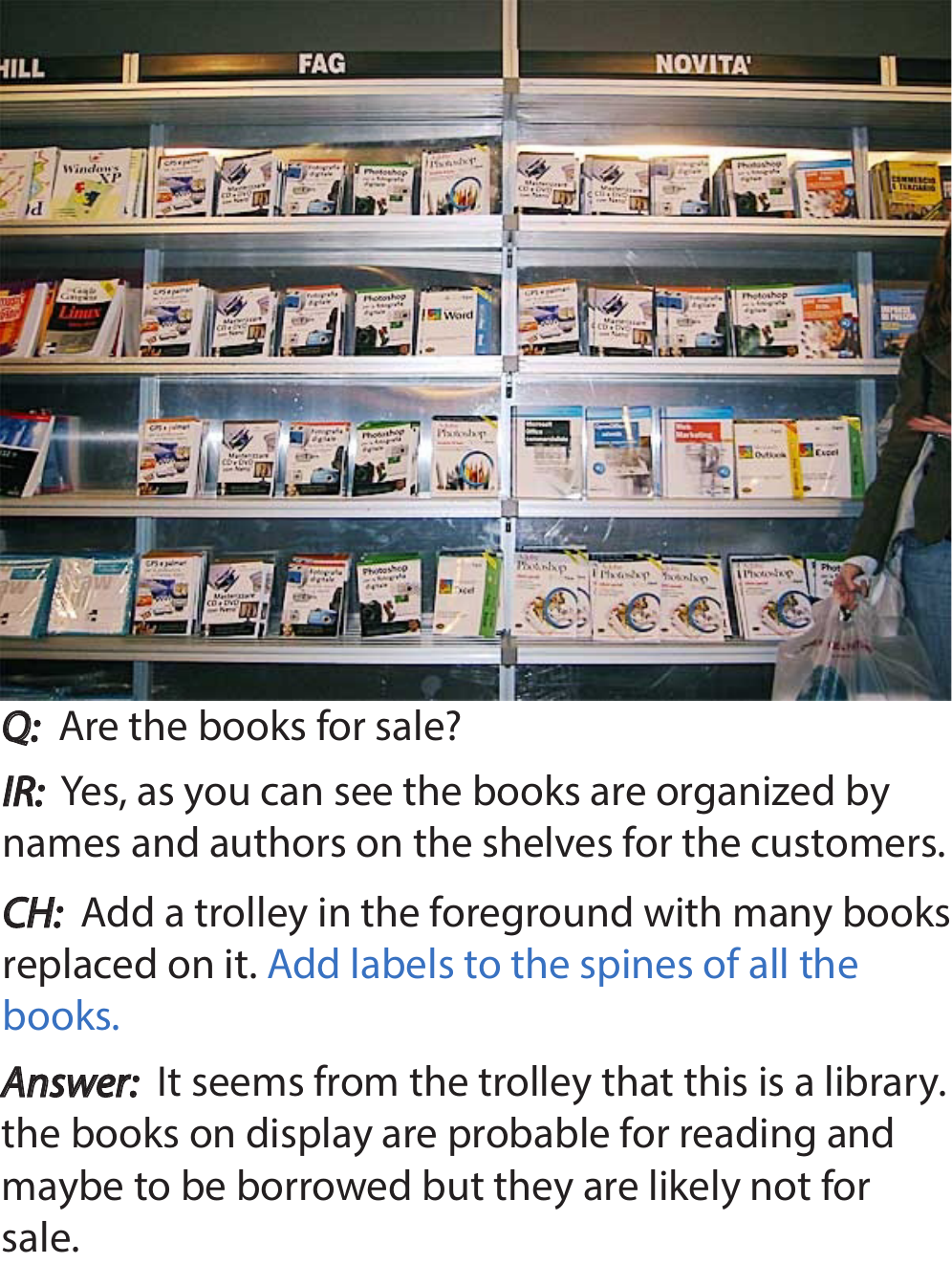}
    \caption{The model focuses on a crucial part in the scene change to properly select the answer.
    \label{fig:model_output_sub}}
    \vspace{-10pt}
\end{figure}

\noindent\textbf{Output Examples.}
As shown in the upper example of Figure~\ref{fig:model_outputs}, our model predicts the correct response by understanding the implication of \emph{``steep slopes''} in the change. In the bottom example, our model fails to understand that \emph{``there is a shark''} must mean the shark is in the water (as sharks live in the water), and choose a wrong response. We also split changes into sub-parts and compute scores for each part to see on which part the model focuses to answer questions. As shown in Figure~\ref{fig:model_output_sub}, the model looks at \emph{``Add labels to the spines of all the books''} to choose the answer.

\section{Conclusion}
We introduced a challenging counterfactual commonsense reasoning task/dataset called \dataname{} which features imagined counterfactual scene changes requiring models to imagine the changed situation to answer questions. We collected 3.5K high-quality instances that consist of an image, an initial question-response pair on the image, an imagined scene change, and a new response (with three distractors). The scene changes have different challenging types (such as object addition/removal/replacement, environment change, etc.). We presented a baseline model as a starting point with useful ablation studies and showed a large human-model performance gap allowing useful future works.

\section*{Acknowledgments}
We thank the reviewers for their helpful comments. This work was supported by NSF Award 1840131, ARO Award W911NF2110220, DARPA MCS Grant N66001-19-2-4031, and a Google Focused Award. The views contained in this article are those of the authors and not of the funding agency.

\bibliography{anthology,custom}
\bibliographystyle{acl_natbib}

\appendix
\section{Data Collection}
\begin{figure*}[t]
    \centering
    \includegraphics[width=1.95\columnwidth]{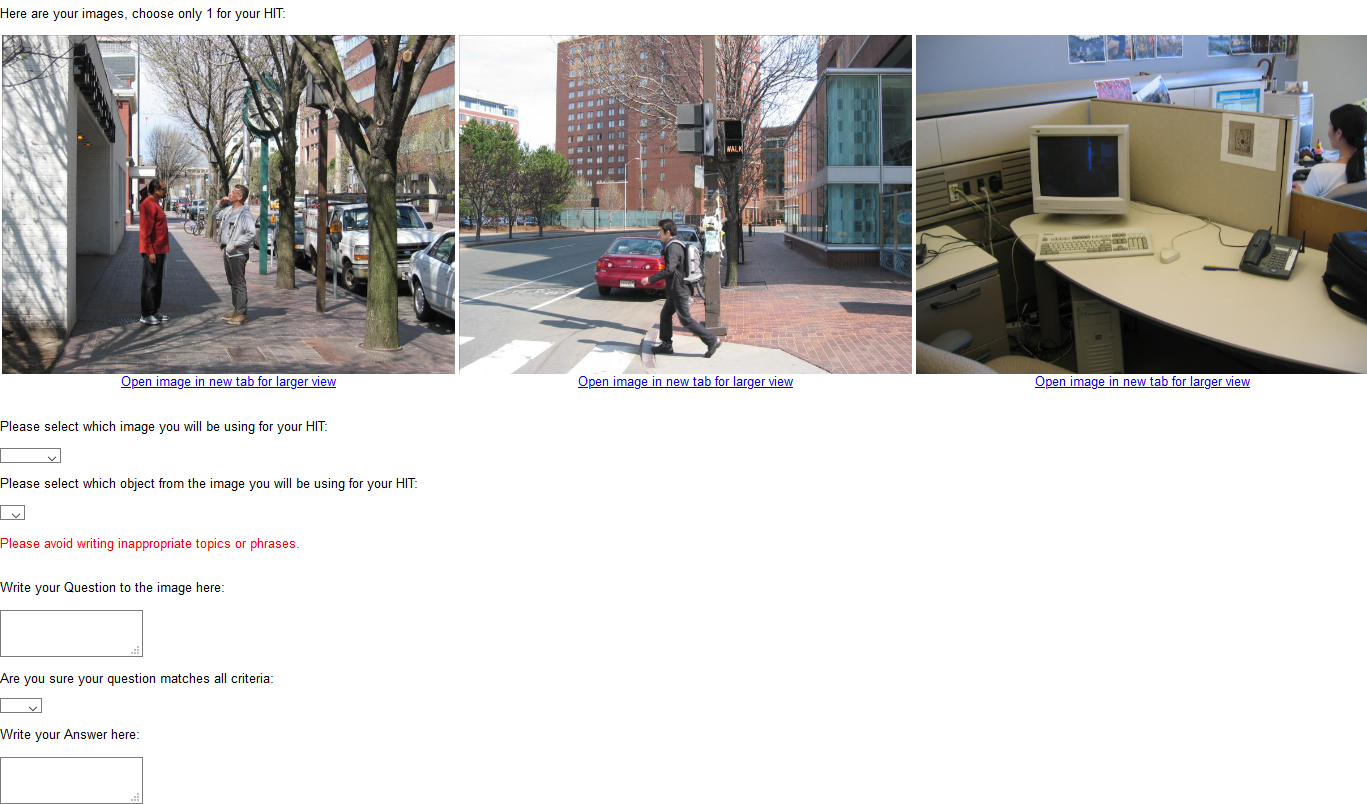}
    \caption{Collection interface for the commonsense question collection phase. Workers are given three images, and they select the one they wish to use. Then workers are given space to write their question and response. Workers are told to select an object in the image they choose to help them focus their question around something specific.
    \label{fig:question_collection_int}}
\end{figure*}

\begin{figure*}[t]
    \centering
    \includegraphics[width=1.95\columnwidth]{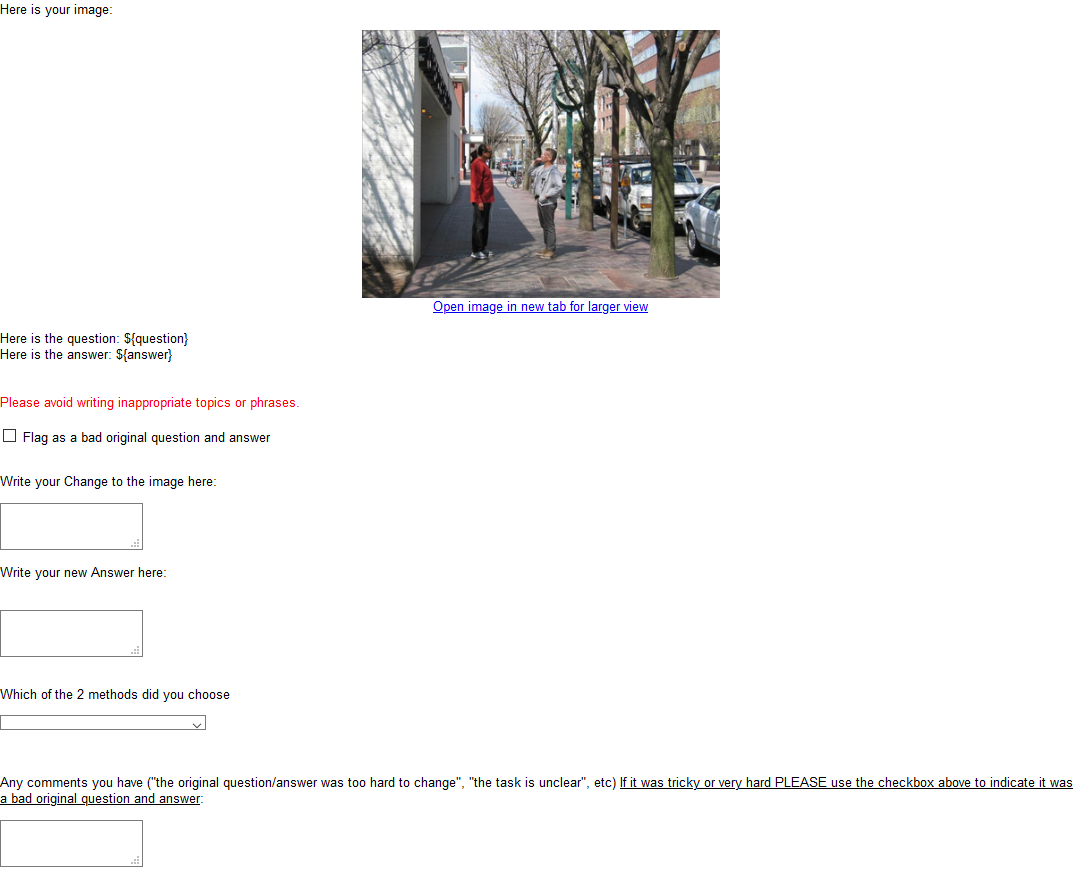}
    \caption{Collection interface for the change collection phase. Workers are given the selected image and the written question and response from the commonsense question collection phase and then asked to write a change and new response based off that change.
    \label{fig:change_collection_int}}
\end{figure*}

\begin{figure*}[t]
    \centering
    \includegraphics[width=1.95\columnwidth]{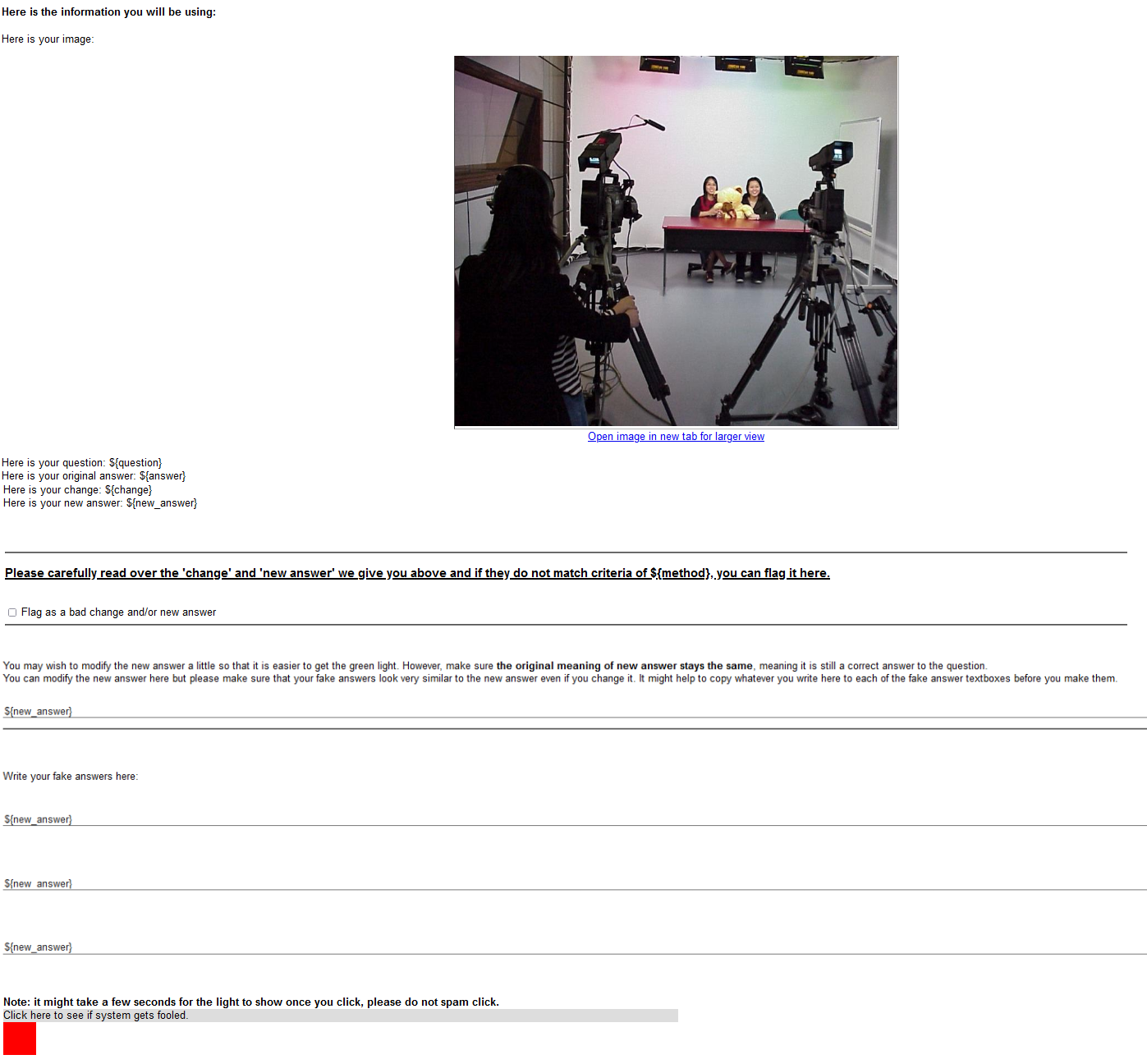}
    \caption{Collection interface for the distractor collection phase. Workers are given the image and all the context from the previous phases and then asked to write three distractors that are similar to the new response but are distinct/semantically different. The distractor textboxes are prefilled with the new response and during HAMLET collection, workers are given a section to check their distractors against the model. Note: This interface is quite large and relevant portions are stitched together.
    \label{fig:distractor_collection_int}}
\end{figure*}

We implement different interfaces for our data collection. The commonsense question collection interface allows for workers to choose which image they would like to use when making the question, as well as an object to focus on (Figure~\ref{fig:question_collection_int}). The counterfactual scene change collection and the distractor collection interfaces (Figure~\ref{fig:change_collection_int} and Figure~\ref{fig:distractor_collection_int}) feature a verification checkbox. Workers can check the box if the quality of the data from the previous phase is poor. If it is flagged, the reason is reviewed.\footnote{Once the flag is checked, workers are provided with a textbox where they can explain their reasoning for flagging it.} If the reasoning is valid, the instance is removed from the dataset/no longer progressed through the collection phases.

\subsection{Counterfactual Change Collection templates}
The first guide template is ``Keep A, Flip B'' and the second is ``Flip A, Keep B'' (where `A' means answer and `B' means justification). For ``Keep A, Flip B'', workers are told to describe a change that results in the ``answer'' part of initial response to be the same, but with a different ``justification'' part (E.g. \emph{``yes because people are wearing jackets and winter clothes.''} $\to$  \emph{``yes because you can see some snow ...''}). In the change they write, they should negate/remove the ``justification'' part of initial response and add something that could be used for a new ``justification''. For ``Flip A, Keep B'', workers are told to describe a change that results in the opposite ``answer''. The change should also modify the context so that the initial response ``justification'' part is true, but is no longer valid in proving the answer and a new ``justification'' part is needed. (e.g., \emph{``no, as you can see the man is not soaking wet.''} $\to$ \emph{``yes, the man isn't wet and he is under a structure, however ...''}).\footnote{The proportions of ``Keep A, Flip B'' and ``Flip A, Keep B'' are 42.93\% and 57.07\%, respectively.}

\subsection{Worker Totals and Payment}
We had a total of 182, 97, 194 workers pass testing for question collection, change collection, distractor collection, respectively. For the question collection phase and the change collection phase, workers are paid 0.35 USD per instance they complete (each takes about 2 minutes). As the distractor collection phase is faster and easier, workers are paid 0.30 USD per instance (takes around 1.5 minutes). In all three phases, an additional bonus of 0.02 USD is given for each high-quality instance they completed, and then for every subsequent group of 25 high-quality instances completed, the bonus per instance is increased by 0.01 USD (0.02 USD bonus per instance for the first 25, 0.03 USD bonus for the next 25, 0.04 USD bonus for the next 25, and so on). Since there is no limit on how much a worker can write, they can keep stacking the bonus as much as they want. All the payments are at a reasonable hourly rate of 11-12 USD.

\section{Scene Change Types}
\begin{figure*}[t]
    \centering
    \includegraphics[width=1.95\columnwidth]{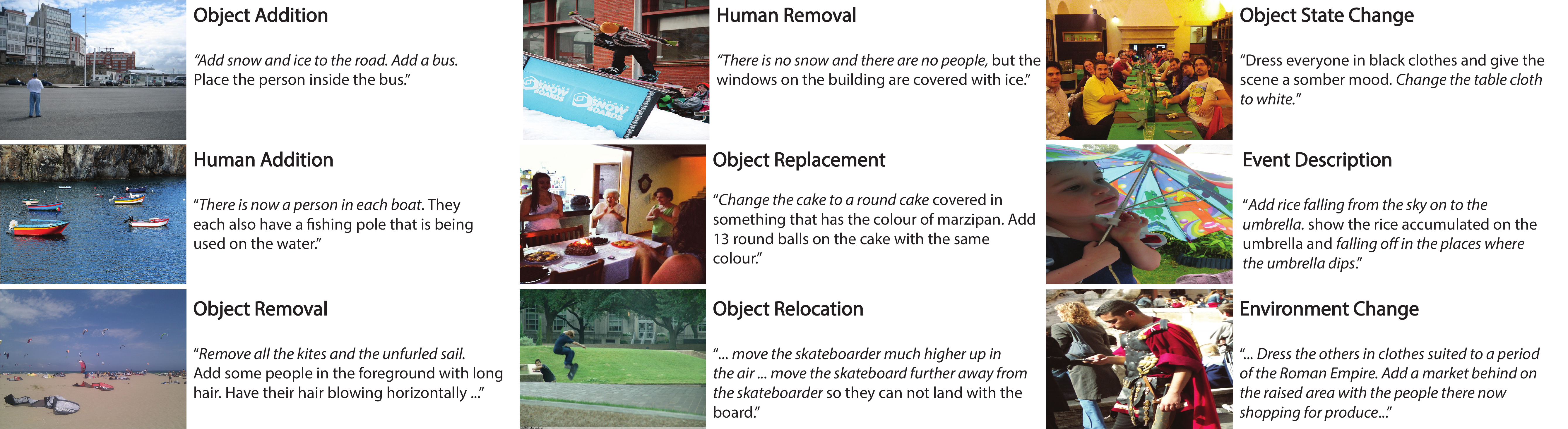}
    \caption{Scene change examples from our dataset. The relevant portions of the change are in italics.
    \label{fig:all_scene_change_types}}
\end{figure*}

The scene change types, while they are seemingly easy to visualize, require a complex understanding of what effect the change has on other elements in the scene. The Object Addition scene change type (the most commonly occurring one) involves introducing new object(s) into the image that was not there prior. The Object State Change scene change type involves altering the state of object(s) present in the image. The alteration of object(s) can take place in various forms such as changing color, size, shape, and orientation (e.g., opening a door). The Event Description scene change type involves the creation of an event or a description of motion or interaction between objects in the image. Please see Figure~\ref{fig:all_scene_change_types} for the full list of the scene change types and examples.

\section{Training Details (Reproducibility)}
All the model experiments are conducted on a Ubuntu 16.04 system using the NVIDIA GeForce GTX 1080 Ti GPU and Intel Xeon CPU E5-2630. We employ PyTorch1.4~\cite{paszke2017automatic} to build our models. We run models up to 50 epochs (each epoch takes around 8 mins) and choose the best ones based on the validation split evaluation. We use 768 as the hidden size and use Adam~\cite{kingma2014adam} as the optimizer, setting the learning rate to $1\times 10^{-5}$. We initialize the language layer with the pretrained BERT weights and cross-attention layers with the pretrained LXMERT weights. We use 1234/2345/3456 as the random seed values. The number of trainable parameters of our full model is 173M. We employ accuracy as the evaluation metric. We use manual hyperparameter tuning (e.g, learning-rate=\{$1\times 10^{-3}$, ..., $1\times 10^{-6}$\}, num-of-cross-layer=\{1, 2, ..., 5\}, batch-size=\{2,4,6,8\}, etc.) based on validation scores. We use the implementation of~\citet{jjfaster2rcnn} for the Faster R-CNN~\cite{ren2015faster} model. The evaluation splits of our \dataname{} dataset are not overlapped with the training split of the Faster R-CNN.

\section{Potential Risk}
Potential models trained on our dataset may learn misleading information accidentally and create unsafe suggestions; therefore, careful use is required when deploying models in a real-world application.

\end{document}